\documentclass{article} 
\usepackage{iclr2018_conference,times}
\usepackage{hyperref}
\usepackage{url}

\usepackage{amssymb} 
\usepackage{color} 
\usepackage{mathrsfs} 
\usepackage{bm} 
\usepackage{graphicx} 
\usepackage{enumitem}
\usepackage{amsmath}

\usepackage[nottoc]{tocbibind}
\usepackage[utf8]{inputenc} 
\usepackage[T1]{fontenc}    
\usepackage{multirow}
\usepackage{booktabs}       
\usepackage{nicefrac}       
\usepackage{microtype}      
\usepackage{subcaption}
\usepackage{xcolor}
\usepackage{array}
\usepackage[T1]{fontenc}
\usepackage{fix-cm}
\usepackage{amsthm} 
\usepackage{amsopn} 

\newcolumntype{K}[1]{>{\centering\let\newline\\\arraybackslash\hspace{0pt}}m{#1}}

\usepackage{tikz}

\DeclareMathOperator*{\argmax}{arg\,max}
\DeclareMathOperator*{\argmin}{arg\,min}

\newtheorem{innercustomgeneric}{\customgenericname}
\providecommand{\customgenericname}{}
\newcommand{\newcustomtheorem}[2]{%
	\newenvironment{#1}[1]
	{%
		\renewcommand\customgenericname{\textbf{#2}}%
		\renewcommand\theinnercustomgeneric{\textbf{##1}}%
		\innercustomgeneric
	}
	{\endinnercustomgeneric}
}

\newcustomtheorem{customthm}{Theorem}
\newcustomtheorem{customlemma}{Lemma}
\newcustomtheorem{customcorr}{Corollary}

\graphicspath{ {figs/} }

\title{HyperNetworks with statistical filtering for defending adversarial examples}

\author{Zhun Sun, Mete Ozay, Takayuki Okatani \\
Graduate School of Information Sciences\\
Tohoku University\\
6-3-09 Aoba, Aramaki-aza Aoba-ku, Sendai, 980-8579, Japan  \\
\texttt{\{sun, mozay, okatani\}@vision.is.tohoku.ac.jp} \\
}

\iclrfinalcopy 

\begin{document}

\maketitle

\begin{abstract}
Deep learning algorithms have been known to be vulnerable to adversarial perturbations in various tasks such as image classification. This problem was addressed by employing several defense methods for detection and rejection of particular types of attacks. However, training and manipulating networks according to particular defense schemes increases computational complexity of the learning algorithms. In this work, we propose a simple yet effective method to improve robustness of convolutional neural networks (CNNs) to adversarial attacks by using data dependent adaptive convolution kernels. To this end, we propose a new type of HyperNetwork in order to employ statistical properties of input data and features for computation of statistical adaptive maps. Then, we filter convolution weights of CNNs with the learned statistical maps to compute dynamic kernels. Thereby, weights and kernels are collectively optimized for learning of image classification models robust to adversarial attacks without employment of additional target detection and rejection algorithms. 

We empirically demonstrate that the proposed method enables CNNs to spontaneously defend against different types of attacks, e.g. attacks generated by Gaussian noise, fast gradient sign methods \citep{goodfellow2014explaining} and a black-box attack \citep{narodytska2016simple}.
\end{abstract}


\section{Introduction}
Deep convolutional neural networks are powerful and popular algorithms that achieve state-of-the-art performance in various computer vision tasks, such as object recognition. Despite the advances made by the recent architectures \citep{krizhevsky2012imagenet, simonyan2014very, szegedy2015going, he2016deep}, they are discovered to be fragile to small but carefully directed perturbations of images \citep{szegedy2013intriguing}, such that the targeted images can be classified to incorrect categories with high confidence, while humans are still able to  correctly classify the attacked images, being undisturbed or even unaware of the perturbations. The vulnerability of these networks to these, so called \textit{adversarial examples}, may lead to undesirable consequences in safety- and security-critical applications. \citet{papernot2017practical} provide an example of misclassification of traffic signs which could be a significant threat for autonomous driving systems that employ deep learning algorithms. Various adversarial attack methods for neural networks have been studied in numerous works. The majority of attack methods can be catalogued in three groups. 
\begin{enumerate}[leftmargin=0.5cm,itemindent=.5cm,labelwidth=\itemindent,labelsep=0cm,align=left]
    \item Methods which use unspecific statistical noise: In this group, input images are perturbed using unspecific statistical noise, e.g. Gaussian noise, salt and pepper noise and blurring. Since shape and parameters of distribution functions that are used to generate noise are not determined, it is usually not easy to obtain a highly confident misclassification results with imperceptible perturbations \citep{szegedy2013intriguing}.
    
    \item Gradient based attack methods: They are used to generate high confidence imperceptible adversarial examples within few steps or one-shot gradient based noise. Some examples of the methods considered in this group are (Iterative) Fast Gradient Sign Method \citep{goodfellow2014explaining, kurakin2016adversarial}, L-BFGS \citep{tabacof2016exploring}, Jacobian-based Saliency Map \citep{papernot2016limitations} and DeepFool \citep{moosavi2016deepfool}. These methods require a white-box environment in order to make attacks. In other words, the full network architecture and weights are required to be accessible in order to obtain gradients towards input images.
    
    \item Black-box attack methods. These methods assume that only the output of the networks can be accessed. Substitute networks \citep{papernot2017practical} and greedy search of noisy pixels \citep{narodytska2016simple} are considered in this group. It is worth mentioning that, methods such as transferring adversarial examples from another network, which is optimized with a sufficient part or the whole training datasets, are not considered as a \textit{genuine} black-box method.
\end{enumerate}

In this work, inspired by the recent works \citep{de2016dynamic, ha2016hypernetworks} that construct neural networks with data dependent weights, we propose a simple yet effective method to train CNNs by improving their robustness to adversarial perturbations. Our main idea is to adaptively filter convolution weights of CNNs by using statistical properties of input data and features. Concretely, we propose a HyperNetwork to compute statistical adaptive maps using these statistical properties (mean and variance) of input data and features for each input channel. Then, we obtain data dependent kernels for convolution operations by computing Hadamard (element-wise) product of computed maps and convolution weights. Our main contributions can be summarized as follows:
\begin{enumerate}[leftmargin=0.5cm,itemindent=.5cm,labelwidth=\itemindent,labelsep=0cm,align=left]
    \item We propose a new type of CNN architecture that employ HyperNetworks to dynamically generate data dependent convolution kernels with statistical properties of input data and features.
    \item We empirically verify the robustness of our proposed models using large scale vision dataset, and demonstrate that their robustness is improved without using additional aforementioned computationally complex defense methods or spending effort to generate adversarial examples for training.
\end{enumerate}

\subsection*{Related works on defense and HyperNetwork}
Several defense methods have been proposed in the last decade, e.g. evolving uncertainty during training \citep{papernot2016distillation}, training with adversarial examples \citep{jin2015robust, zheng2016improving, tramer2017ensemble}, and training a smarter conjugate network for detecting adversarial perturbations \citep{metzen2017detecting}. 

The most intuitive approach is to employ adversarial examples during training phase. \citet{goodfellow2014explaining} propose to augment the training set with adversarial examples. That is, they simultaneously minimize the loss for original examples and the adversarial ones that are generated according to the aforementioned fast gradient sign method based on current weights. \citet{tramer2017ensemble} further ensemble the training data with adversarial examples produced from pre-trained models. They suggest that training with adversarial examples produced from the model being trained will lead to a \textit{degenerate minima}, where the model is still undefended, except that its gradient points into a non-adversarial direction. \citet{zheng2016improving} propose to append a stability term to the objective function, which regularize the model to produce similar outputs for original examples and their perturbed versions, without considering the classification loss of perturbed examples. This approach is experimentally shown to be able to maintain or improve state-of-the-art performance on the original task.

\citet{papernot2016distillation} propose an approach based on defensively distilling knowledge from a conjugate network that shares the same structure. Precisely, the conjugate network is trained first with \textit{hard} (binary) labels to prepare \textit{soft} ($0-1$ probability) labels, a temperature $\mathit{T}$ is further employed to control the uncertainty during the distillation. Although facing the risk of degraded performance, this type of defense is able to prohibit gradient based attacks by shrinking the magnitude of gradients with respect to the input image. \citet{metzen2017detecting} propose a method to augment deep neural networks with a conjugate network which is trained on the binary classification task of distinguishing genuine data from data containing adversarial perturbations. They show empirically that adversarial perturbations can be detected surprisingly well even though they are quasi-imperceptible to humans.

However these methods are usually cumbersome to carry out, either require more resources such as memories for the conjugate network, or longer training period due to the harder convergence properties caused by adversarial examples \citep{tramer2017ensemble}. In addition, their performance for clean examples may decrease. Moreover, these defense methods are not proposed to be cross-domain in general, and most of them are designed using prior information on incoming attacks, and have less or none robustness toward other types of attacks. 

Recently \citet{de2016dynamic} propose a type of neural network using weights generated dynamically conditioned on an input (\textit{Dynamic Filter}) with a small network. This type of architecture is able to increase flexibility of neural network without an excessive increase in the number of model parameters. They have empirically verified that a wide variety of filtering operations, such as local spatial transformations, as well as selective (de)blurring or adaptive feature extraction can be learned in this way. \citet{ha2016hypernetworks} further generalize the architecture with \textit{HyperNetworks}, and propose the static HyperNetworks as a weights factorization approach for deep convolutional networks. Although proved to be powerful, this architecture has not been used in large scale vision tasks due to the high dimension of weights in the recent state-of-the-art convolutional neural network architectures.

\section{Our Approach}
\subsection{Adversarial Examples in Deep Learning}
We define adversarial examples as follows. Suppose that we are given a deep convolutional neural network optimized for standard object classification tasks $\bm{\mathit{p}}_{\bm{I}} = \mathcal{F}_\mathcal{W} (\bm{I})$, where $\bm{\mathit{p}} \in \mathbb{R}^\mathcal{C}$ is a vector of probability values, and $\mathcal{W}$ is the set of weights used at all layers of the network. Then the prediction of the network can be expressed as 
$\mathit{y}_p = \argmax_{\mathit{c}} (\bm{\mathit{p}}_{\bm{I}})$,
where $\mathit{p}$ is the predicted probability for each class $\mathit{c} \in \mathcal{C}$. The ground truth label of $\bm{I}$ is denoted by $\mathit{y}$. An adversarial example is denoted by ${\bm{I}}^{\mathit{adv}}$, whose prediction label is $\mathit{y}_p \neq \mathit{y}$, while it is kept relatively close to the original example $\bm{I}$, $\mathfrak{D}(\bm{I}, \bm{I}^{\mathit{adv}}) < \epsilon$, where $\mathfrak{D}$ is a chosen measure such as the $\mathit{l}_2$ distance. The desired output either can be  trivial that only makes the network mis-classify $\bm{I}^{\mathit{adv}}$, or can be targeted to a special class $\mathit{y}_t$ such that $\mathit{y}_p = \mathit{y}_t$ with a high confidence.

\begin{figure}[t]
\centering
    \begin{minipage}[t]{0.6\linewidth}
    \centering
        \includegraphics[width=\textwidth]{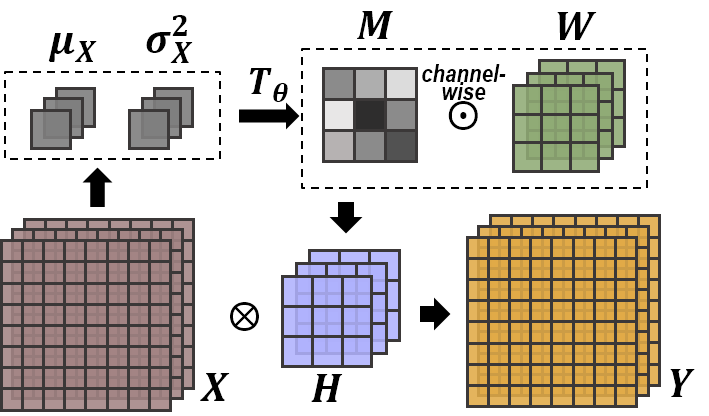}
    \end{minipage}
\caption{An overview of the proposed Hyper-Convolution layer at which statistical properties of data are used.}
\vspace{-0.5cm}
\label{fig:struct}
\end{figure}

\subsection{Statistical Hyper-Convolution}
An $L$ layer convolutional neural network $\mathcal{F}_\mathcal{W}$ can be expressed as a series transformations each of which applies a spatial convolution operation $G^l$ by $ \bm{Y}^{l} = \sigma^l(\mathit{G}^{l}(\mathbf{W}^{l}, \bm{X}^{l}))$ at the $l^{th}$ layer, where  $\sigma$ is a functional module such as activation, normalization or soft-max, and $\bm{X}^{0} = \bm{I}, \bm{X}^{l+1} = \bm{Y}^{l}$. We use $\mathbf{W}^{l}$ to denote the corresponding weights that are employed at each layer,  and $\mathcal{W} = \{{\bm{W}}^l \}_{l=1}^L$ is the set of weights used in the network. In the later discussions, we omit the superscripts of the layer index $l$ for simplicity. 

Vanilla CNNs carry out convolution between feature maps (or an input image) and stationary kernels by $\mathit{G}(\mathbf{W}, \bm{X}) = \mathbf{W} \otimes \bm{X}$, where $\mathbf{W} \in \mathbb{R}^{C \times D \times S \times S}$, $\bm{X} \in \mathbb{R}^{D \times H \times W}$, $C$ is the number of output channels, $D$ is the number of input channels, and  $S$ is the size of kernels. Our main idea is to adaptively filter convolution weights of CNNs using a HyperNetwork \citep{ha2016hypernetworks} ${\bm{T}}_{\bm{\theta}}$ that is parameterized by $\bm{\theta}$. For this purpose, the network ${\bm{T}}_{\bm{\theta}}$ receives the channel-wise mean and standard deviation of input $\bm{X}$ as its input, and outputs a map $\bm{M} \in \mathbb{R}^{S \times S}$, which can be computed by $\bm{M} = {\bm{T}}_{\bm{\theta}} (\bm{\mu}_{\bm{X}}, \bm{\sigma}_{\bm{X}})$, where $\bm{\mu}_{\bm{X}}, \bm{\sigma}_{\bm{X}} \in \mathbb{R}^{D}$. Then, we compute Hadamard (element-wise) product ($\odot$) over each $S \times S$ sub-kernels of the convolution weight $\bm{W}$, and compute the map $\bm{H}_{C, D} = \mathbf{W}_{C, D} \odot \bm{M}$. Finally, we employ the obtained adaptive convolution kernels to perform the convolution operation, by $\mathit{G} (\mathbf{H}, \bm{X})= \bm{H} \otimes \bm{X}$. Figure~\ref{fig:struct} provides an overview of this procedure. We choose the network ${\bm{T}}_{\bm{\theta}}$ to be an ordinary two-layer neural network which employs $\mathit{l}_2$ regularization, and which has $D/2$ neurons in the hidden layer. ReLU is used as the activation function at the hidden layer, while the output layer employs sigmoid such that the elements of $\bm{M}$ take values in $(0, 1)$. We use $\mathcal{T}_{\bm{\Theta}} = \{ {\bm{T}}^l_{\bm{\theta}_l} \}_{l=1}^L $ to denote the set of ${\bm{T}}_{\bm{\theta}}$ employed at each layer, and $\mathcal{F}_{\mathcal{W}, \mathcal{T}}$ to denote the convolutional neural network equipped with SHC. 

The parameters $\theta$ computed at each layer are updated simultaneously with $\mathcal{W}$ using SGD with momentum after receiving the gradient $\frac{\partial J}{\partial \theta}$ by $\frac{\partial J}{\partial {\bm{Y}}} \cdot \frac{\partial {\bm{Y}}}{\partial {\bm{H}}} \cdot \frac{\partial {\bm{H}}}{\partial {\bm{M}}} \cdot \frac{\partial {\bm{M}}}{\partial \theta}$. It is worthy mentioning that, HyperNetworks also back-propagate errors (or gradients) from upper to bottom layers by $\frac{\partial {\bm{Y}}}{\partial {\bm{M}}} \cdot \frac{\partial {\bm{M}}}{\partial {\bm{X}}}$. We call this back-propagation route the HyperNetwork route, and the ordinary back-propagation route passing through the convolution operation the  Convolution route. The total back-propagated error can be computed as the summation of errors propagating from both routes at each layer.

\subsection{Attacks used in this work}
\noindent \textbf{Gaussian Noise} Gaussian noise is the most commonly used attack type. In this paper, we employ a truncated version of Gaussian noise that restricts change of pixels within $[-\epsilon, \epsilon]$ by
\begin{equation}
\label{gnoise}
    \bm{I}^{\mathit{adv}} = \mathrm{Clip} [~\bm{I} + \mathcal{N}(\bm{0}, \bm{\sigma}),~\epsilon~].
\end{equation}

\noindent \textbf{Fast Gradient Sign} \citet{goodfellow2014explaining} proposed a white-box method to find perturbations by
\begin{equation}
\label{fgsm}
    \bm{I}^{\mathit{adv}} = \bm{I} + \epsilon \cdot \mathrm{Sign} (\bm{\nabla}_{\bm{I}} J(\bm{I}, \mathit{y})),
\end{equation}
where $J(\bm{I}, \mathit{y})$ is a loss function such as $\mathit{y}_p $ for the trivial case ($\mathit{y}_p = \mathit{y}$), or standard categorical cross-entropy (CCE) loss with the targeted class $\mathrm{CCE}(\mathit{y}_t,  \mathbf{p}_{\bm{I}})$. In this paper, we employ 
$\mathit{y}_t = \argmin_{\mathit{c}} (\bm{\mathit{p}}_{\bm{I}})$ 
as the targeted class.

This method can be extended to an iterative style \citep{kurakin2016adversarial}, where a smaller bound $\alpha < \epsilon$ is used for each step, and the output of the previous iteration will be clipped to apply changes in $[-\epsilon, \epsilon]$.
\begin{equation}
\label{ifgsm}
    \bm{I}^{\mathit{adv}}_0 = \bm{I}, ~~~\bm{I}^{\mathit{adv}}_{n+1} = \mathrm{Clip} [~\bm{I} + \alpha \cdot \mathrm{Sign} (\bm{\nabla}_{\bm{I}} J(\bm{I}, \mathit{y})),~\epsilon~].
\end{equation}

\citet{tramer2017ensemble} proposed a simple yet powerful novel attack that first applies a small random perturbation to an input, before finding the optimal perturbation under a first-order approximation, which can be formulated by
\begin{equation}
\label{rfgsm}
    \bm{I}' = \bm{I} + \alpha \cdot \mathrm{Sign}(\mathcal{N}(\bm{0}, \bm{I})), ~~~\bm{I}^{\mathit{adv}} = \bm{I}' + \epsilon \cdot \mathrm{Sign} (\bm{\nabla}_{\bm{I}'} J(\bm{I}', \mathit{y})).
\end{equation}

\noindent \textbf{LocSerachAdv} \citet{narodytska2016simple} proposed an iterative approach that can efficiently locate a small set of pixels, without using any gradient information, which leads to mis-classification by a deep neural network when it is perturbed. The algorithm targets to push the true label $\mathit{y}$ below the $\mathit{k}^{th}$ variable of the probability vector $\bm{\mathit{p}}_{\bm{I}}$. Briefly, it takes the output from the previous step (which is a clean image for the first step), and generates different perturbed versions, using a semi-random method according to the location of perturbed pixels in the previous step. Then, it performs a greedy search to select the pixels that fool the network most, and construct a new (perturbed) image. Although it has a chance of failure, this type of attack is still indefensible by far, since the applied perturbations do not depend on the architecture of the target network nor how the network is optimized.

\section{Experimental Analysis}
\subsection{Measurement of model robustness}
\begin{figure}[t]
\centering
    \begin{minipage}[t]{1\linewidth}
    \centering
        \begin{minipage}[t]{0.16\linewidth}
        \centering
            $\mathrm{Clean}$ 
        \end{minipage}
        \begin{minipage}[t]{0.16\linewidth}
        \centering
            $\mathrm{RCD\texttt{=}1.13}$ 
        \end{minipage}
        \begin{minipage}[t]{0.16\linewidth}
        \centering
            $\mathrm{RCD\texttt{=}3.08}$ 
        \end{minipage}
        \begin{minipage}[t]{0.16\linewidth}
        \centering
            $\mathrm{RCD\texttt{=}4.76}$ 
        \end{minipage}
        \begin{minipage}[t]{0.16\linewidth}
        \centering
            $\mathrm{RCD\texttt{=}5.85}$ 
        \end{minipage}
        \begin{minipage}[t]{0.16\linewidth}
        \centering
            $\mathrm{RCD\texttt{=}6.54}$ 
        \end{minipage}
    \end{minipage}\vspace{0.05cm}
    \begin{minipage}[t]{1\linewidth}
    \centering
        \begin{minipage}[t]{0.16\linewidth}
        \centering
            \includegraphics[width=\textwidth]{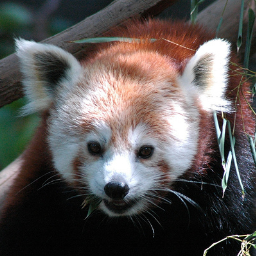}
            \subcaption*{\scriptsize $\sigma\texttt{=} 0.00, \bm{\textit{p}}_\textit{y}\texttt{=} 0.94$}
        \end{minipage}
        \begin{minipage}[t]{0.16\linewidth}
        \centering
            \includegraphics[width=\textwidth]{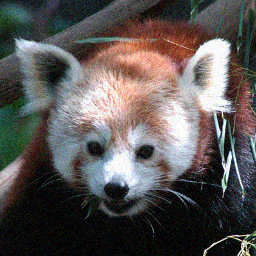}
            \subcaption*{\scriptsize $\sigma\texttt{=} 0.05, \bm{\textit{p}}_\textit{y}\texttt{=} 0.77$}
        \end{minipage}
        \begin{minipage}[t]{0.16\linewidth}
        \centering
            \includegraphics[width=\textwidth]{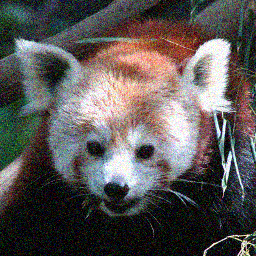}
            \subcaption*{\scriptsize $\sigma\texttt{=} 0.10, \bm{\textit{p}}_\textit{y}\texttt{=} 0.33$}
        \end{minipage}
        \begin{minipage}[t]{0.16\linewidth}
        \centering
            \includegraphics[width=\textwidth]{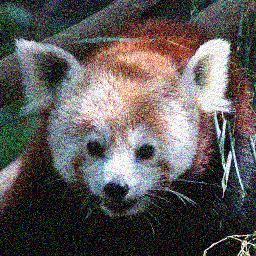}
            \subcaption*{\scriptsize $\sigma\texttt{=} 0.15, \bm{\textit{p}}_\textit{y}\texttt{=} 0.11$}
        \end{minipage}
        \begin{minipage}[t]{0.16\linewidth}
        \centering
            \includegraphics[width=\textwidth]{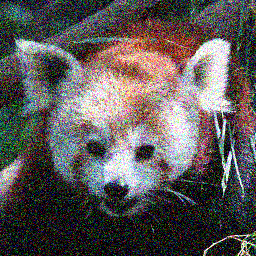}
            \subcaption*{\scriptsize $\sigma\texttt{=} 0.20, \bm{\textit{p}}_\textit{y}\texttt{=} 0.05$}
        \end{minipage}
        \begin{minipage}[t]{0.16\linewidth}
        \centering
            \includegraphics[width=\textwidth]{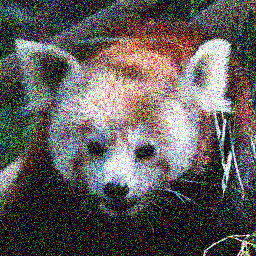}
            \subcaption*{\scriptsize $\sigma\texttt{=} 0.25, \bm{\textit{p}}_\textit{y}\texttt{=} 0.01$}
        \end{minipage}
    \end{minipage}
\caption{An example of the RCD score obtained using different levels of Gaussian noise attacks.}
\vspace{-0.5cm}
\label{fig:rcd}
\end{figure}

In the experimental evaluation of the method, we introduce a new method, called Relative Confidence Diminution (RCD) score, to measure its robustness to attacks in addition to classification accuracy. More precisely, RCD score is used to compute the difference between the Relative Confidence (RC), which can be formulated as the log ratio of confidence for the correct label and the variance of the noisy labels by
\begin{equation}
\label{RCD}
    \mathrm{RCD}_{\mathcal{F}}(\bm{I}^{adv}, \bm{I}) = \mathrm{RC}_{\mathcal{F}}(\bm{I}) - \mathrm{RC}_{\mathcal{F}}(\bm{I}^{adv}) = 
    \mathrm{log} \frac{{\bm{\mathit{p}}_{\bm{I}^{adv}, \mathit{y}}}} {{\mathrm{SD}[ \bm{\mathit{p}}_{\bm{I}^{adv}, \mathit{k}} ]_{\mathit{k} \neq \mathit{y}}}} - \mathrm{log} \frac{{\bm{\mathit{p}}_{\bm{I}, \mathit{y}}}} {{\mathrm{SD}[ \bm{\mathit{p}}_{\bm{I}, \mathit{k}} ]_{\mathit{k} \neq \mathit{y}}}},
\end{equation}
where $\mathrm{SD}$ stands for Standard Deviation. Intuitively, RCD score can be considered as a difference in ``signal noise ratio'' between predictions, and provide an intuition on how much the model is affected, when the adversarial perturbation is not strong enough to lead a misclassification. An adversarial perturbation that increases the confidence of an incorrect label will result a higher $\mathrm{SD}[ \bm{\mathit{p}}_{\bm{I}, \mathit{k}} ]_{\mathit{k} \neq \mathit{y}}$, thus we will obtain a larger RCD score. Figure~\ref{fig:rcd} shows an example of RCD score obtained  by employment of Gaussian noises with different $\sigma$.

\subsection{Experiment configurations and results}
In this section, we employ three sets of different models for evaluation: ResNet-18, ResNest-18-SHC; ResNet-50, ResNet-50-SHC (the SHC is only employed in convolution layers with $3 \times 3$ kernel size) \citep{he2016deep}; a smaller prototype 10-layer plain network (denoted as Plain-10 and Plain-10-SHC). The Plain-10 network has the same number of channels for output feature maps as ResNet, but only has one convolution block (two $3 \times 3$ convolution layers) at each resolution. These models are optimized by classifying 1000 classes of ILSVRC-2012 dataset \citep{ILSVRC15}, using the same learning scheme provided in \citet{he2016deep}. The training and validation images are re-scaled within range $[0, 1]$, all the hyper-parameters of the attacks follow this range. During testing, the validation images are first re-sized to $299 \times 299$, then the single center crop of $256 \times 256$ is fed into the networks. We employ 3 types of aforementioned attacks to generate adversarial examples. In this work, we put stress on the robustness of the networks other than the perceivability of perturbations. Therefore, beside generating imperceptible perturbations, we also introduce sets of parameters that are able to create \textit{heavier} perturbations on the images, which may be able to be perceived by human in some cases.

\noindent \textbf{Gaussian noise} In this test, we evaluate the robustness of network towards random Gaussian noise on the whole validation set of ILSVRC-2012, and the noisy examples are generated using \eqref{gnoise}. We perform the tests using three different standard deviation ($\sigma$) of noise and the results are reported in Table~\ref{table:gnoise}. It can be seen that the performance of networks equipped with SHC is boosted in all the cases, and their robustness toward Gaussian noise is kept the same with the Plain-10 network, and it is improved by $\sim 2\%$ for ResNet-18 and ResNet-50.

\setlength{\tabcolsep}{5.2pt}
\begin{table}[t]
  \caption{Analysis of robustness of the models to random Gaussian noise. The accuracy and the RCD scores are averaged over all 50,000 validation examples.}
  \centering
  \begin{tabular}{ccccccccc}
    \toprule
    \multirow{2}[2]{*}{\textbf{Network}} & \textbf{clean} & 
                \multicolumn{2}{c}{$\bm{\sigma \texttt{=} 0.1}$} & 
                \multicolumn{2}{c}{$\bm{\sigma \texttt{=} 0.2}$} & 
                \multicolumn{2}{c}{$\bm{\sigma \texttt{=} 0.4}$} \\
    \cmidrule(lr){2-2}\cmidrule(lr){3-4}\cmidrule(lr){5-6}\cmidrule(lr){7-8}
      & \textbf{Acc.}(\%) & 
        \textbf{Acc.}(\%) & \textbf{RCD} & 
        \textbf{Acc.}(\%) & \textbf{RCD} & 
        \textbf{Acc.}(\%) & \textbf{RCD} \\
    \midrule
    Plain-10        & 60.74 & 49.49 & 0.89 & 43.88 & 1.24 & 31.23 & 2.11 \\
    Plain-10-SHC    & 64.32 & 53.25 & 0.92 & 47.63 & 1.36 & 34.02 & 2.44 \\
    \midrule
    ResNet-18       & 71.16 & 62.49 & 0.92 & 58.43 & 1.29 & 48.41 & 2.19 \\
    ResNet-18-SHC   & 73.48 & 64.61 & 0.95 & 61.08 & 1.32 & 52.58 & 2.18 \\
    \midrule
    ResNet-50       & 76.67 & 69.36 & 1.00 & 65.79 & 1.41 & 57.55 & 2.44 \\
    ResNet-50-SHC   & 77.79 & 71.12 & 0.92 & 67.89 & 1.31 & 61.02 & 1.96 \\
    \bottomrule
  \end{tabular}
  \vspace{-0.5cm}
  \label{table:gnoise}
\end{table}

\noindent \textbf{Gradient based attack} In this test, we evaluate the robustness of network to Fast Gradient Sign Method and it's variants. For each comparison pair, we select 5,000 validation images that are correctly classified by both reference networks and SHC networks with high confidence ($>0.9$), and create adversarial examples according to the equations (\ref{fgsm}), (\ref{ifgsm}) and (\ref{rfgsm}). For the original FGSM, we employ the parameter $\epsilon = 0.02$, and then we further employ another $\epsilon = 0.1$ (denoted by FSGM-Heavy) to examine the performance of proposed method under strong adversarial perturbations. For the Iterative FGSM attack, we employ $\epsilon=0.02$ and $\alpha=0.001$ with 5 epochs. For the Rand FGSM attack, we employ $\epsilon=0.03125$ and $\alpha=0.03125$ which perturb at most $8/256$ of the pixel values. The results of averaged accuracy and RCD are given in Table~\ref{table:fgsm}. Obviously, even without utilizing any type of defensive techniques during training, the proposed SHC networks obtain robustness to the gradient based attacks regardless of the methods and their strength. Meanwhile, it can be seen from the RCD scores that the gradient based attacks are better at confusing the networks than pure noise, and the effectiveness of different methods is different for different models. For instance, the RAND+FGSM is more effective in attacking ResNet-50 than I-FGSM for similar RCD scores but at a lower accuracy, while its effectiveness is on par with I-FGSM for ResNet-50-SHC. Moreover, for vanilla CNNs, the robustness is highly related with the depth (or training difficulty) of architectures, which suggests that the robustness is related to the back-propagation of attack gradients. On the other hand, the SHC based ResNet-50 performs worse than ResNet-18 in 3 of 4 situations, and we will further discuss on this observation in Section~\ref{sec:discuss2}.

\begin{table}[t]
  \caption{Analysis of robustness of the models against fast gradient sign method and its variants. We attack 5,000 images that are initially correctly classified with high confidence ($>0.9$), the accuracy and RCD score are averaged over the corresponding 5,000 adversarial examples.}
  \centering
  \begin{tabular}{ccccccccc}
    \toprule
    \multirow{2}[2]{*}{\textbf{Network}} & 
                \multicolumn{2}{c}{\textbf{FGSM}} & 
                \multicolumn{2}{c}{\textbf{FGSM-Heavy}} & 
                \multicolumn{2}{c}{\textbf{I-FGSM}} & 
                \multicolumn{2}{c}{\textbf{RAND+FGSM}} \\
                
    \cmidrule(lr){2-3}\cmidrule(lr){4-5}\cmidrule(lr){6-7}\cmidrule(lr){8-9}
    & \textbf{Acc.}(\%) & \textbf{RCD} & 
      \textbf{Acc.}(\%) & \textbf{RCD} & 
      \textbf{Acc.}(\%) & \textbf{RCD} & 
      \textbf{Acc.}(\%) & \textbf{RCD} \\
    \midrule
    Plain-10        &  3.06 & 10.54 &  0.94 & 11.51 &  1.90 & 14.11 &  0.40 & 12.57 \\
    Plain-10-SHC    & 16.66 &  9.44 &  8.38 & 10.58 & 14.14 & 10.91 &  6.98 & 11.77 \\
    \midrule
    ResNet-18       & 10.54 &  9.78 &  3.54 & 11.67 &  6.18 & 12.48 &  1.94 & 11.46 \\
    ResNet-18-SHC   & 44.34 &  7.53 & 23.64 &  9.68 & 37.56 &  8.74 & 28.16 &  9.21 \\
    \midrule
    ResNet-50       & 28.46 &  8.84 & 20.02 &  9.84 & 16.78 & 10.68 &  9.94 & 10.72 \\
    ResNet-50-SHC   & 42.40 &  7.88 & 31.16 &  8.99 & 28.40 &  9.57 & 27.38 &  9.52 \\
    \bottomrule
  \end{tabular}
  \vspace{-0.5cm}
  \label{table:fgsm}
\end{table}

\begin{figure}[t]
\centering
    \begin{minipage}[t]{1\linewidth}
    \centering
        \begin{minipage}[t]{0.47\linewidth}\centering
            \includegraphics[width=\textwidth]{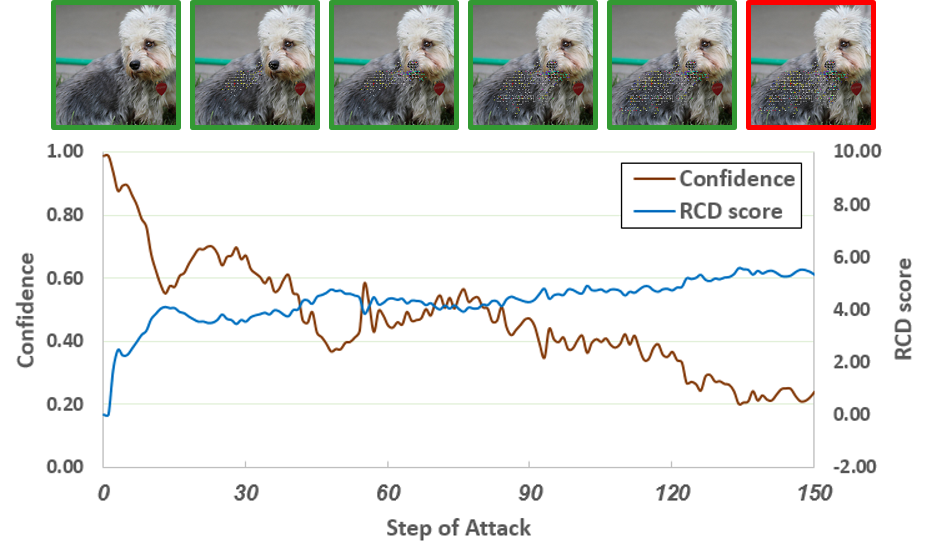}\subcaption{}
        \end{minipage}
        \begin{minipage}[t]{0.47\linewidth}\centering
            \includegraphics[width=\textwidth]{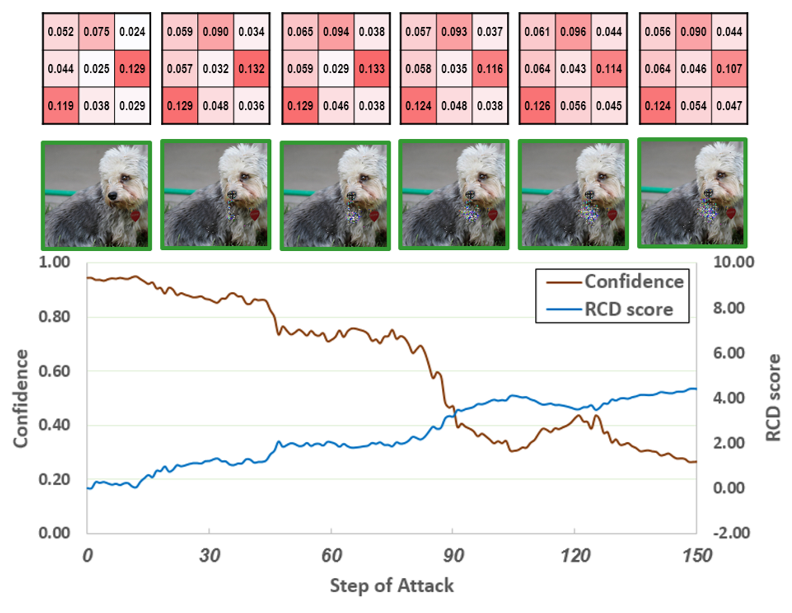}\subcaption{}
        \end{minipage}
    \end{minipage}
\vspace{-0.1cm}
\caption{An example of an LSA attack towards (a) Vanilla ResNet-50, and (b) ResNet-50-SHC. The change of confidence of correct class and RCD scores along with attack steps are depicted by brown and blue curves, respectively. The images with ``best'' perturbations are shown at the top of the figure. We mark the images that are correctly classified with green frames, and the images that are mis-classified with red frames. We also provide the obtained $\bm{M}$ at the last convolution layer of ResNet-50-SHC. In this case, the confidence for vanilla ResNet-50 drops fast at the beginning of the attack, and results in a failure in defence of the attack. The proposed method maintains a high confidence until step 90, and correctly classified the adversarial examples at step 150.}
\vspace{-0.3cm}
\label{fig:lsa}
\end{figure}

\noindent \textbf{LocSearchAdv} The LocSearchAdv algorithm is employed for examining the robustness of proposed method to black-box attacks. We use most of the configurations proposed in the original paper, with an increased number of pixels perturbed at each step. That is, 10 pixels are selected to be perturbed for a non-stop 150 steps, thus a total of 1,500 pixels are selected and perturbed. The target of perturbations is set to decrease the confidence of the correct class regardless of the fact that the current image has been mis-classified already. We select 500 images that are correctly classified with high confidence for both networks, and record the change of average accuracy and RCD at different steps, these results are provided in Table~\ref{table:lsa}. Obviously, the proposed method performs better within the first dozens of steps of the LSA attack. It usually takes about 60 more steps for SHC models to obtain similar RCD scores as the reference model obtains. An example of how the image changes together with its confidence and RCD is shown in Figure~\ref{fig:lsa}. It is worth noticing that, given enough attempts, the method is able to successfully generate adversarial examples for most images though, it usually takes more attempts to fool the proposed method.

\begin{table}[t]
  \caption{Model robustness against LSA attack. We select 500 images that are initially correctly classified with high confidence ($>0.9$), and we apply the LSA algorithm to search the corresponding adversarial examples for 150 steps. We report the averaged RCD over attack steps.}
  \centering
  \begin{tabular}{ccccccccccc}
    \toprule
    \multirow{2}[2]{*}{\textbf{Network}} & 
        \multicolumn{2}{c}{\textbf{Step 30}} & 
        \multicolumn{2}{c}{\textbf{Step 60}} & 
        \multicolumn{2}{c}{\textbf{Step 90}} & 
        \multicolumn{2}{c}{\textbf{Step 120}} & 
        \multicolumn{2}{c}{\textbf{Step 150}} \\
    \cmidrule(lr){2-3}\cmidrule(lr){4-5}\cmidrule(lr){6-7}\cmidrule(lr){8-9}\cmidrule(lr){10-11}
    & \textbf{Acc.} & \textbf{RCD} & \textbf{Acc.} & \textbf{RCD} & \textbf{Acc.} & \textbf{RCD} & \textbf{Acc.} & \textbf{RCD} & \textbf{Acc.} & \textbf{RCD}\\
    \midrule
    Plain-10        & 87.2 & 1.40 & 77.8 & 2.03 & 69.0 & 2.46 & 63.0 & 2.82 & 57.8 & 3.14 \\
    Plain-10-SHC    & 97.8 & 0.75 & 93.2 & 1.19 & 87.0 & 1.55 & 82.2 & 1.84 & 78.4 & 2.14 \\
    \midrule
    ResNet-18       & 81.6 & 1.49 & 71.2 & 2.15 & 62.4 & 2.58 & 54.8 & 2.99 & 51.8 & 3.24 \\
    ResNet-18-SHC   & 93.4 & 0.75 & 88.9 & 1.11 & 84.0 & 1.43 & 80.2 & 1.66 & 75.2 & 1.85 \\
    \midrule
    ResNet-50       & 77.6 & 1.61 & 66.4 & 2.28 & 59.4 & 2.72 & 55.2 & 3.23 & 47.6 & 3.52 \\
    ResNet-50-SHC   & 90.4 & 0.80 & 84.0 & 1.22 & 77.0 & 1.55 & 72.4 & 1.87 & 67.6 & 2.15 \\
    \bottomrule
  \end{tabular}
  \vspace{-0.1cm}
  \label{table:lsa}
\end{table}

\begin{table}[t]
  \caption{An experimental analysis of robustness of the model to adversarial examples generated using gradients from different routes. We attack 5,000 images that are initially correctly classified with high confidence ($>0.9$). The accuracy and RCD score are averaged over the corresponding 5,000 adversarial examples.}
  \centering
  \begin{tabular}{ccccccccccc}
    \toprule
    \multirow{2}[2]{*}{\textbf{Network}} & 
    \multicolumn{2}{c}{\textbf{RAND-FGSM}} & 
    \multirow{2}[2]{*}{\textbf{Network}} & 
    \multicolumn{2}{c}{\textbf{RAND-FGSM}} & \\
    \cmidrule(lr){2-3}\cmidrule(lr){5-6}
    & \textbf{Acc.(\%)} & \textbf{RCD} & & \textbf{Acc.(\%)} & \textbf{RCD}\\
    \cmidrule(lr){1-3}\cmidrule(lr){4-6}
    ResNet-18        &  1.94 & 11.46 &    ResNet-18-SHC-T     & 98.5 &  1.18 \\
    ResNet-18-SHC    & 28.16 &  9.21 &    ResNet-18-SHC-W     &  2.6 & 13.29 \\
    \bottomrule
  \end{tabular}
  \vspace{-0.1cm}
  \label{table:discuss}
\end{table}

\begin{figure}[t]
\centering
    \begin{minipage}[t]{0.9\linewidth}\centering
        \begin{minipage}[t]{0.21\columnwidth}\centering
            \includegraphics[width=\textwidth]{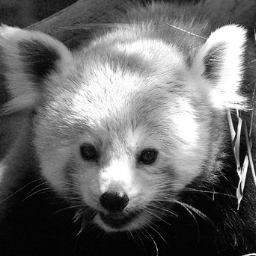}\subcaption{}
        \end{minipage}
        \hspace{0.1cm}
        \begin{minipage}[t]{0.21\columnwidth}\centering
            \includegraphics[width=\textwidth]{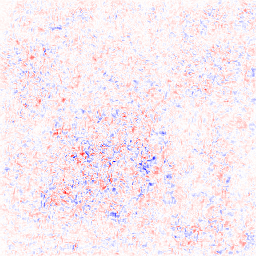}\subcaption{}
        \end{minipage}
        \hspace{0.1cm}
        \begin{minipage}[t]{0.21\columnwidth}\centering
            \includegraphics[width=\textwidth]{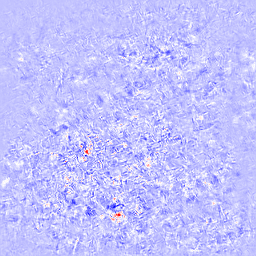}\subcaption{}
        \end{minipage}
        \hspace{0.1cm}
        \begin{minipage}[t]{0.21\columnwidth}\centering
            \includegraphics[width=\textwidth]{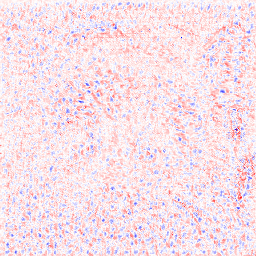}\subcaption{}
        \end{minipage}
    \end{minipage}
\vspace{-0.1cm}
\caption{The gradients $\nabla_{\bm{I}}{J}$ obtained from different routes. (a) Red channel of the input image, (b) full gradients obtained from both routes, (c) gradients obtained from the convolution route, (d) gradients obtained from the HyperNetwork route.}
\vspace{-0.5cm}
\label{fig:discuss}
\end{figure}


        

\section{Discussion}
\subsection{Effect of black-box attacks to class decision boundary}
Considering a binary image classification problem, a class decision boundary of a deep neural network (DNN) can be defined as a hypersurface that partitions the input space into two target classes. Once the optimization of the DNN is finished, the decision boundary is computed, and the robustness of the DNN is determined by the properties of the decision boundary \citep{fawzi2016robustness, fawzi2017geometric}. In this section, we employ the concept of the \textit{quasi decision boundary} to examine the observed robustness of an SHC network, when a network is attacked by a black-box greedy search algorithm such as LocSearchAdv.

Given an SHC network $\mathcal{F}_{\mathcal{W}, \mathcal{T}}$ and an input image $\bm{I}$, we define a neighbourhood of an input as the set of images that differ from the input by small perturbations by  
\[
\mathcal{B}_{\bm{I}} = \{ {\bm{J}} | \mathfrak{D}({\bm{I}}, {\bm{J}}) < \epsilon_1, \mathfrak{D}( {\bm{T}}_{\theta}(\bm{I}), {\bm{T}}_{\theta}(\bm{J}) ) < \epsilon_2 \},
\]
while the change of the output of the HyperNetwork with respect to the perturbations can be ignored. We assume that, the perturbed images from LSA at each step are in the neighbourhood of the current input, since the perturbation of several pixels can barely result in changes of statistics. Figure~\ref{fig:lsa}(b) provides an example of the change of $\bm{M}$ due to perturbations at every 30 steps. Now we state that a \textit{quasi decision boundary}, is the decision boundary of a sub-network $\mathcal{F}_{\mathcal{H}}$, where $\mathcal{H} = \{ {\bm{H}} \}^L_{l=1}$ is the set of convolution kernels that are computed by the HyperNetwork with respect to an input image $\bm{I}$ and its neighbourhood. 

When the problem is restricted to classification of the given image and the images belonging to its neighbourhood, this type of \textit{quasi decision boundary} can be considered as a good approximation of the true decision boundary. However, it will be broken once the target is out from the neighbourhood. We could consider the greedy search algorithm of LSA as follows: at each step, a set of random perturbed inputs \{$\hat{\bm{I}}\} \subset \mathcal{B}_{\bm{I}}$ is used to estimate the strength of perturbations. Then, the \textit{best} perturbations are accumulated to generate the adversarial image $\bm{I}'$. However as the progress goes on, the strength of perturbations estimated in former steps will become untrustworthy, since the image $\bm{I}'$ \textit{is more likely to} have left the neighbourhood of former inputs as a consequence of the accumulated perturbations. Thus the attack progress will be halted or even reversed due to the unreliability until it finds the perturbations that breaks the true decision boundary of $\mathcal{F}_{\mathcal{W}, \mathcal{T}}$.

\vspace{-0.1cm}
\subsection{Effect of HyperNetworks against white-box attacks}
\vspace{-0.1cm}
\label{sec:discuss2}
The results in Table~\ref{table:fgsm} appears that the ResNet-50-SHC performs worse towards most gradient base methods, compared to ResNet-18-SHC , regardless of the better accuracy it obtained using clean images and the images perturbed by Gaussian noise. Meanwhile we notice that, for vanilla CNN models, the robustness is related to the training difficulty and the gradient propagation of architectures. Thus in order to analyze the robustness of SHC model, we further carry out a set of experiments to separately attack the sub-network $\mathcal{F}_{\mathcal{H}}$ and HyperNetworks $\mathcal{T}_{\theta}$, respectively. More precisely, we \textit{block} the gradients that pass through $\mathcal{T}_{\theta}$ (the HyperNetworks route) when attacking $\mathcal{F}_{\mathcal{H}}$, and vice versa. Note that these procedures are only employed in generating adversarial examples with respect to the \textit{partially back-propagated} attack gradients, while the generated examples are still fed into the whole SHC-network for evaluation. Figure~\ref{fig:discuss} shows an example of the gradients $\nabla_{\bm{I}}J$ obtained from different routes. 

We employ the RAND-FGSM attack and report the results in Table~\ref{table:discuss}. It can be observed that the proposed method is as easy to be fooled as a vanilla CNN, if we only consider the gradients passing through its sub-network $\mathcal{F}_{\mathcal{H}}$ (SHC-W, which can be treated as a vanilla CNN regarding a single input). On the other hand, most of the examples, which only receive the adversarial attack gradients  that are obtained from the HyperNetworks (SHC-T), can be still correctly classified with small decrease in confidence. This observation suggests that employment of the HyperNetworks together with the statistical properties of input data and features is helpful for weakening (dispersing) the adversarial attack gradients.

\vspace{-0.1cm}
\section{Conclusion}
\vspace{-0.1cm}
In this work, we propose a simple yet effective method to improve robustness of convolutional neural networks (CNNs) to adversarial attacks by training CNNs using data dependent adaptive convolution kernels. To this end, we employ HyperNetworks to dynamically generate data dependent convolution kernels with statistical properties of input data and features. The robustness of our proposed method is verified using 3 different types of attack with state-of-the-art CNN models trained on the ILSVRC-2012 dataset. Moreover, the robustness is obtained spontaneously during a normal training progress without losing any performance in the original tasks. This shed light on building practical deep learning systems that focus on the target without a concern of attacker. On the other hand, there still exists uncertainty on the mechanism of the robustness remains to be solved in the future works. Furthermore, designing of network architectures that employ more powerful HyperNetworks with better adversarial robustness is still an open problem.

\bibliography{arxiv}
\bibliographystyle{iclr2018_conference}

\end{document}